\documentclass{article}
\usepackage{ijcai20}

\usepackage{times}
\usepackage{soul}
\usepackage{url}
\usepackage[hidelinks]{hyperref}
\usepackage[utf8]{inputenc}
\usepackage[small]{caption}
\usepackage{graphicx}
\usepackage{amsmath}
\usepackage{amsthm}
\usepackage{booktabs}
\usepackage{algorithm}
\usepackage{algorithmic}
\usepackage{amssymb}
\usepackage{pifont}
\usepackage{makecell}
\usepackage{subcaption}
\usepackage[title]{appendix}
\usepackage{xcolor}
\usepackage[
  separate-uncertainty = true,
  multi-part-units = repeat
]{siunitx}
\title{ Weakly Supervised Few-shot Object Segmentation \\ using Co-Attention with Visual and Semantic Embeddings}

 \makeatletter
 \newcommand{\printfnsymbol}[1]{%
  \textsuperscript{\@fnsymbol{#1}}%
 }
 \makeatother

\author{
 Mennatullah Siam$^{1, 4}$\thanks{equally contributing}\and
 Naren Doraiswamy$^2$\printfnsymbol{1}\and
 Boris N. Oreshkin$^3$\printfnsymbol{1}\and
 Hengshuai Yao$^4$\\\And
 Martin Jagersand$^1$\\
 \affiliations
 $^1$ University of Alberta\\
 $^2$ Indian Institute of Science\\
 $^3$ Element AI\\
 $^4$ HiSilicon, Huawei Research\\
}

\begin{document}

\maketitle

\begin{abstract}
   Significant progress has been made recently in developing few-shot object segmentation methods. Learning is shown to be successful in few-shot segmentation settings, using pixel-level, scribbles and bounding box supervision. This paper takes another approach, i.e., only requiring image-level label for few-shot object segmentation. We propose a novel multi-modal interaction module for few-shot object segmentation that utilizes a co-attention mechanism using both visual and word embedding. Our model using image-level labels achieves 4.8\% improvement over previously proposed image-level few-shot object segmentation. It also outperforms state-of-the-art methods that use weak bounding box supervision on PASCAL-$5^i$. Our results show that few-shot segmentation benefits from utilizing word embeddings, and that we are able to perform few-shot segmentation using stacked joint visual semantic processing with weak image-level labels. We further propose a novel setup, Temporal Object Segmentation for Few-shot Learning (TOSFL) for videos. TOSFL can be used on a variety of public video data such as Youtube-VOS, as demonstrated in both instance-level and category-level TOSFL experiments.

\end{abstract}

\setlength{\textfloatsep}{5pt plus 1.0pt minus 2.0pt}

\section{Introduction}
\begin{figure}[t]
\centering
    \includegraphics[width=0.4\textwidth]{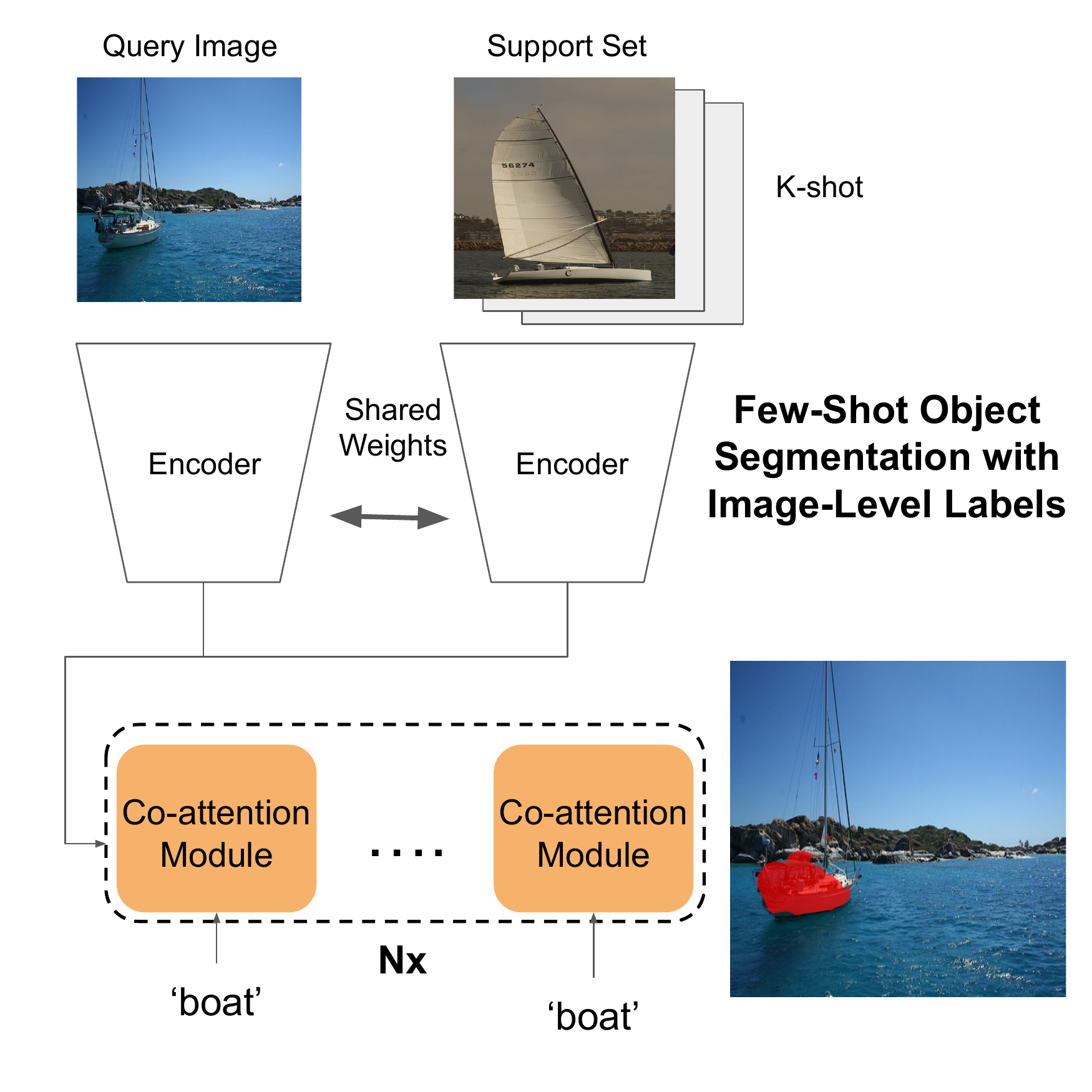}
    \caption{Overview of stacked co-attention to relate the support set and query image using image-level labels. Nx: Co-attention stacked N times. ``K-shot'' refers to using K support images.}
     \label{fig:overview}
\end{figure}

Existing literature in few-shot object segmentation has mainly relied on manually labelled segmentation masks. A few recent works~\cite{rakelly2018conditional,zhang2019canet,wang2019panet} started to conduct experiments using weak annotations such as scribbles or bounding boxes. However, these weak forms of supervision involve more manual work compared to image level labels, which can be collected from text and images publicly available on the web. Limited research has been conducted on using image-level supervision for few-shot segmentation~\cite{raza2019weakly}.

On the other hand, deep semantic segmentation networks are very successful when trained and tested on relatively large-scale manually labelled datasets such as PASCAL-VOC~\cite{everingham2015pascal} and MS-COCO~\cite{lin2014microsoft}. However, the number of object categories they cover is still limited despite the significant sizes of the data used. The limited number of annotated objects with pixel-wise labels included in existing datasets restricts the applicability of deep learning in inherently open-set domains such as robotics~\cite{dehghan2019online,pirk2019online,cordts2016cityscapes}.

In this paper, we propose a multi-modal interaction module to bootstrap the efficiency of weakly supervised few-shot object segmentation by combining the visual input with neural word embeddings. Our method iteratively guides a bi-directional co-attention between the support and the query sets using both visual and neural word embedding inputs, using only image-level supervision as shown in Fig.~\ref{fig:overview}. It outperforms~\cite{raza2019weakly} by 4.8\% and improves over methods that use bounding box supervision~\cite{zhang2019canet,wang2019panet}. We use the term `weakly supervised few-shot' to denote that our method during the inference phase only utilizes few-shot image-level labelled data to guide the class agnostic segmentation. Additionally we propose a novel setup, temporal object segmentation with few-shot learning (TOSFL). The TOSFL setup for video object segmentation generalizes to novel object classes as can be seen in our experiments on Youtube-VOS dataset~\cite{xu2018youtube}. TOSFL only requires image-level labels for the first frames (support images). The query frames are either the consecutive frames (instance-level) or sampled from another sequence having the same object category (category-level). The TOSFL setup is interesting as it is closer to the nature of learning novel objects by human than the strongly supervised static segmentation setup. Our setup relies on the image-level label for the support image to segment different parts from the query image conditioned on the word embeddings of this image-level label. In order to ensure the evaluation for the few-shot method is not biased to a certain category, it is best to split into multiple folds and evaluate on different ones similar to~\cite{shaban2017one}. 

\subsection{Contributions}
\begin{itemize}
    \item We propose a novel few-shot object segmentation algorithm based on a multi-modal interaction module trained using image-level supervision. It relies on a multi-stage attention mechanism and uses both visual and semantic representations.
    \item We propose a novel weakly supervised few-shot video object segmentation setup. It complements the existing few-shot object segmentation benchmarks by considering a practically important use case not covered by previous datasets. Video sequences are provided instead of static images which can simplify the few-shot learning problem.
    \item We conduct a comparative study of different architectures proposed in this paper to solve few-shot object segmentation with image-level supervision. Our method compares favourably against the state-of-the-art methods relying on pixel-level supervision and outperforms the most recent methods using weak annotations~\cite{raza2019weakly,wang2019panet,zhang2019canet}.
\end{itemize}

\section{Related Work}

\begin{figure*}[t]
    \centering
    \includegraphics[width=0.8\textwidth]{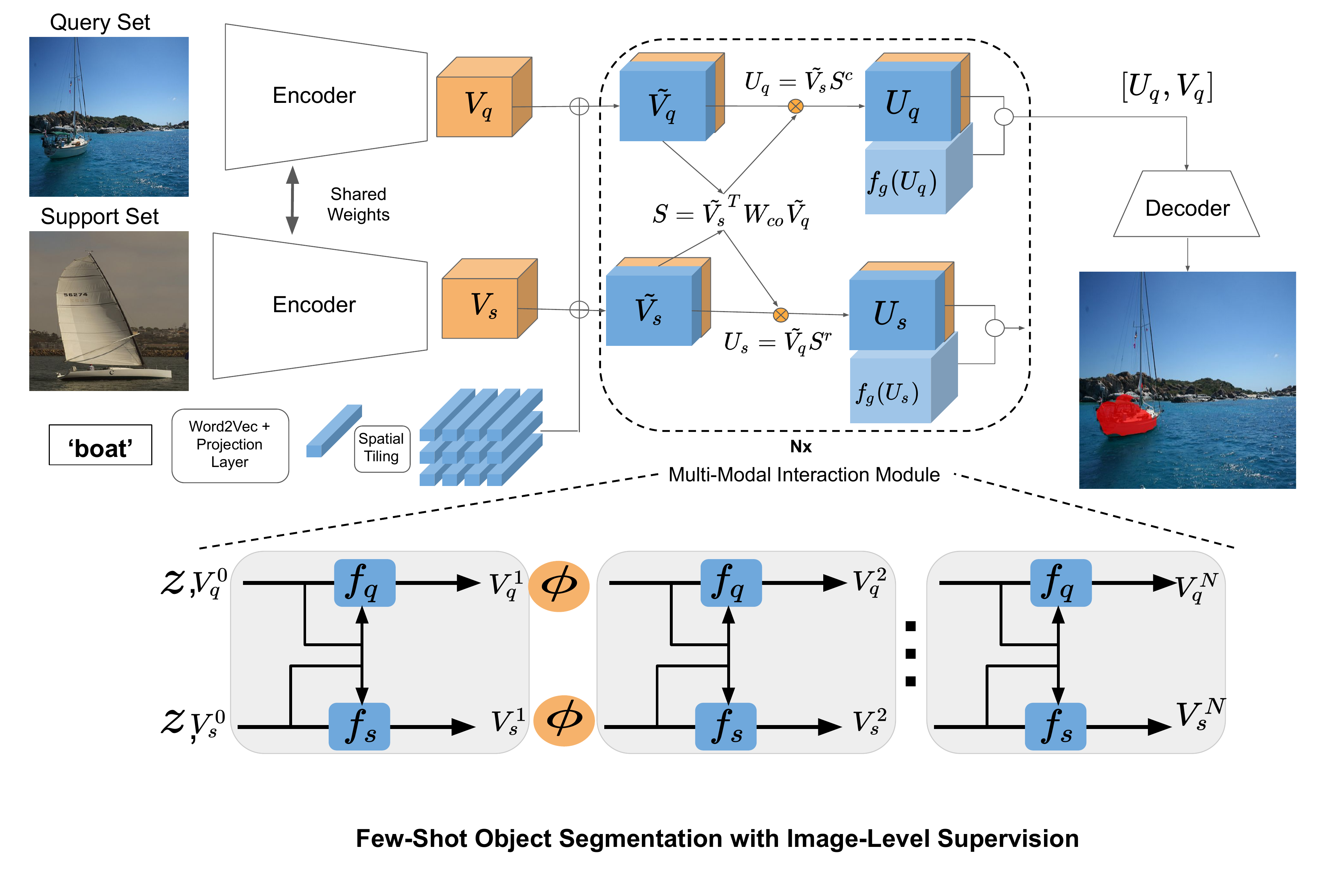}
\caption{Architecture of Few-Shot Object segmentation model with co-attention and overview of the stacked co-attention. The $\oplus$ operator denotes concatenation, $\circ$ denotes element-wise multiplication. Only the decoder and multi-modal interaction module parameters are learned, while the encoder is pretrained on ImageNet.}
\label{fig:detailed}
\end{figure*}

\subsection{Few-shot Object Segmentation}
\cite{shaban2017one} proposed the first few-shot segmentation method using a second branch to predict the final segmentation layer parameters. \cite{rakelly2018conditional} proposed a guidance network for few-shot segmentation where the guidance branch receives the support set image-label pairs. \cite{dong2018few} utilized the second branch to learn prototypes. \cite{zhang2019canet} proposed a few-shot segmentation method based on a dense comparison module with a siamese-like architecture that uses masked average pooling to extract features on the support set, and an iterative optimization module to refine the predictions. \cite{siam2019amp} proposed a method to perform few-shot segmentation using adaptive masked proxies to directly predict the parameters of the novel classes. \cite{zhang2019pyramid} in a more recent work proposed a pyramid graph network which learns attention weights between the support and query sets for further label propagation. \cite{wang2019panet} proposed prototype alignment by performing both support-to-query and query-to-support few-shot segmentation using prototypes. 

The previous literature focused mainly on using strongly labelled pixel-level segmentation masks for the few examples in the support set. It is labour intensive and impractical to provide such annotations for every single novel class, especially in certain robotics applications that require to learn online. A few recent works experimented with weaker annotations based on scribbles and/or bounding boxes~\cite{rakelly2018conditional,zhang2019canet,wang2019panet}. In our opinion, the most promising direction to solve the problem of intense supervision requirements in the few-shot segmentation task, is to use publicly available web data with image-level labels. \cite{raza2019weakly} made a first step in this direction by proposing a weakly supervised method that uses image-level labels. However, the method lags significantly behind other approaches that use strongly labelled data.

\subsection{Attention Mechanisms}
Attention was initially proposed for neural machine translation models~\cite{bahdanau2014neural}. Several approaches were proposed for utilizing attention. \cite{yang2016stacked} proposed a stacked attention network which learns attention maps sequentially on different levels. \cite{lu2016hierarchical} proposed co-attention to solve a visual question and answering task by alternately shifting attention between visual and question representations. \cite{lu2019see} used co-attention in video object segmentation between frames sampled from a video sequence. \cite{hsieh2019one} rely on attention mechanism to perform one-shot object detection. However, they mainly use it to attend to the query image since the given bounding box provides them with the region of interest in the support set image. To the best of our knowledge, this work is the first one to explore the bidirectional attention between support and query sets as a mechanism for solving the few-shot image segmentation task with image-level supervision.

\section{Proposed Method}
The human perception system is inherently multi-modal. Inspired from this and to leverage the learning of new concepts we propose a multi-modal interaction module that embeds semantic conditioning in the visual  processing scheme as shown in Fig.~\ref{fig:detailed}.
The overall model consists of: 
(1) Encoder. (2) Multi-modal Interaction module.  (3) Segmentation Decoder. The multi-modal interaction module is described in detail in this section while the encoder and decoder modules are explained in Section~\ref{sec:imp_details}. We follow a 1-way $k$-shot setting similar to~\cite{shaban2017one}.

\begin{figure*}[t]
\centering
    \begin{subfigure}{.25\textwidth}
    \centering
        \includegraphics[scale=0.35]{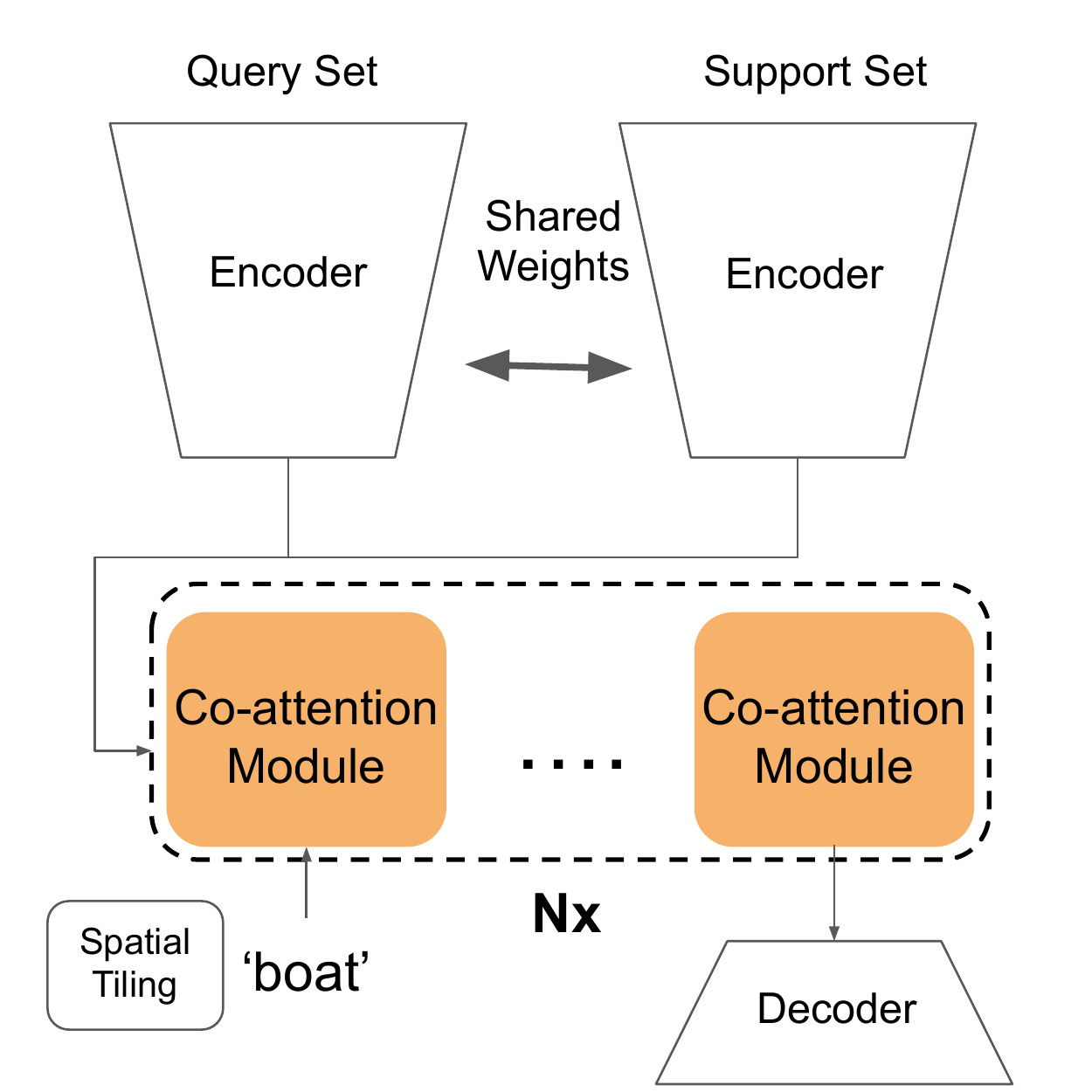}
        \caption{V+S}
    \end{subfigure}%
    \begin{subfigure}{.25\textwidth}
    \centering
        \includegraphics[scale=0.35]{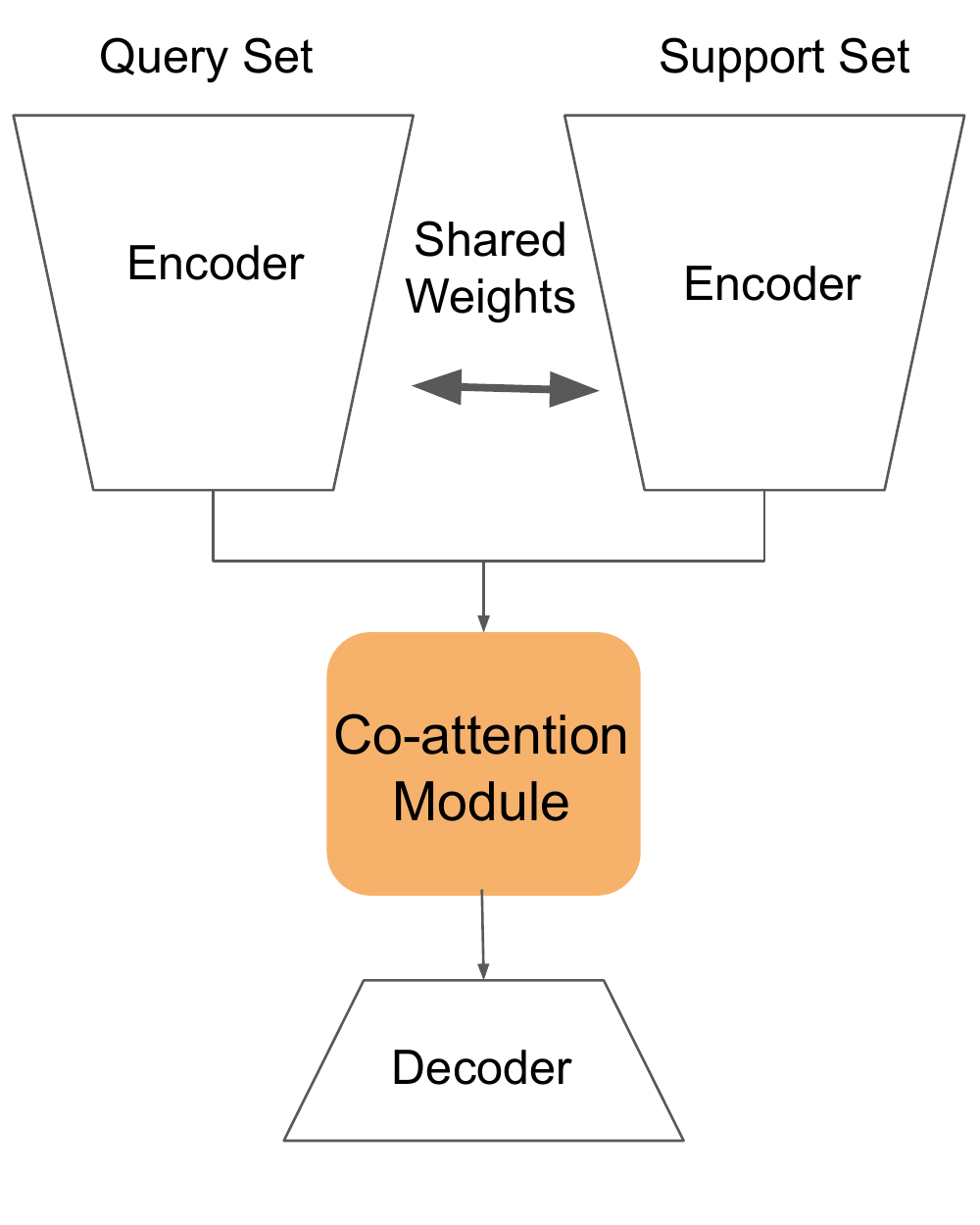}
        \caption{V}
    \end{subfigure}%
    \begin{subfigure}{.25\textwidth}
    \centering
        \includegraphics[scale=0.35]{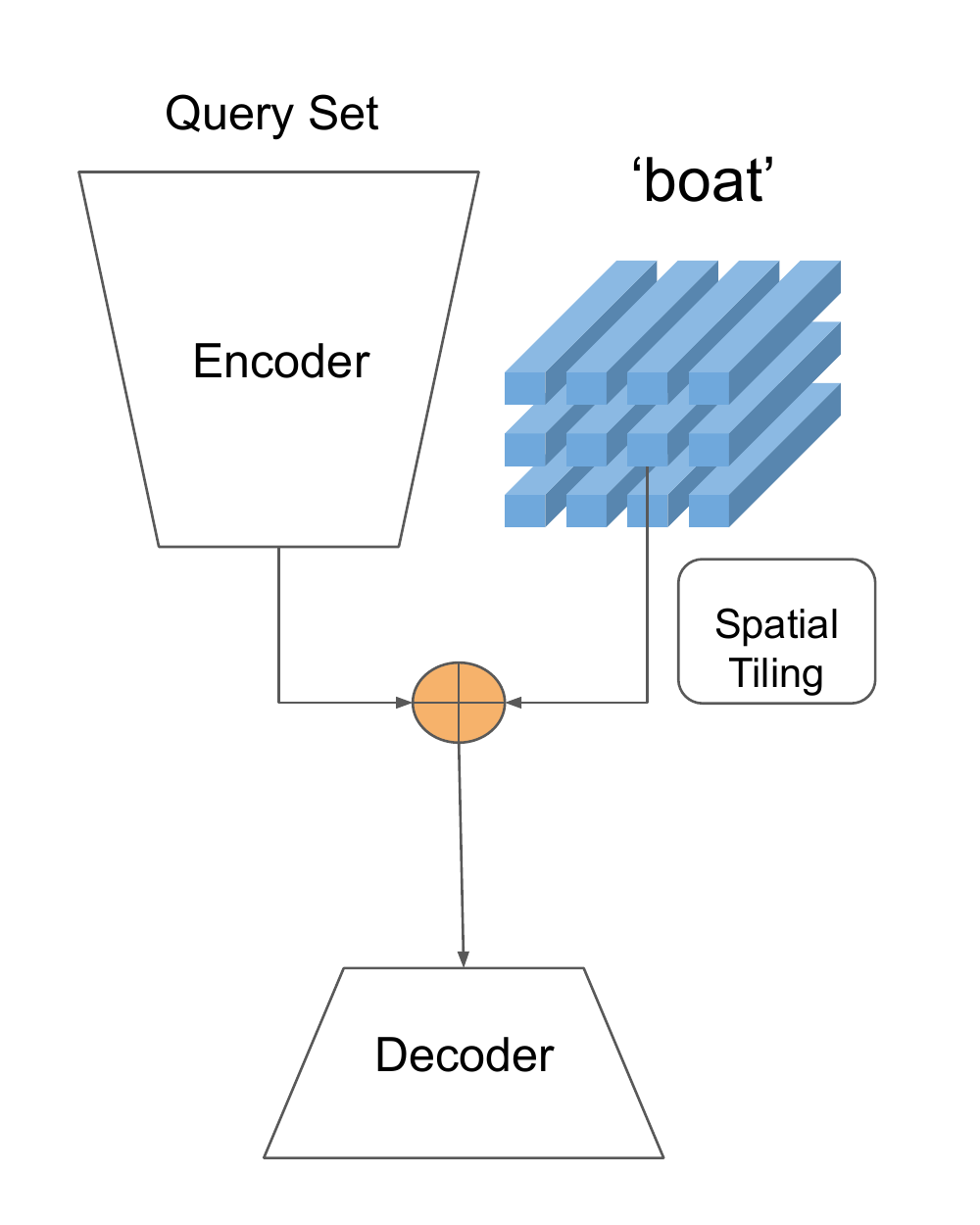} 
        \caption{S} \label{fig:variants:semantic}
    \end{subfigure}%
    \caption{Different variants for image-level labelled few-shot object segmentation. V+S: Stacked Co-Attention with Visual and Semantic representations. V: Co-Attention with Visual features only. S: Conditioning on semantic representation only from word embeddings.}
     \label{fig:variants}
\end{figure*}

\subsection{Multi-Modal Interaction Module} \label{sec:CS2QA}

One of the main challenges in dealing with the image-level annotation in few-shot segmentation is that quite often both support and query images may contain a few salient common objects from different classes. Inferring a good prototype for the object of interest from multi-object support images without relying on pixel-level cues or even bounding boxes becomes particularly challenging. Yet, it is exactly in this situation, that we can expect the semantic word embeddings to be useful at helping to disambiguate the object relationships across support and query images. Below we discuss the technical details behind the implementation of this idea depicted in Fig.~\ref{fig:detailed}. Initially, in a $k$-shot setting, a base network is used to extract features from $i^{th}$ support set image $I_s^i$  and from the query image $I_q$, which we denote as $V_s \in R^{W \times H \times C}$ and $V_q \in R^{W\times H\times C}$. Here $H$ and $W$ denote the height and width of feature maps,  respectively, while $C$ denotes the number of feature channels. Furthermore, a projection layer is used on the semantic word embeddings to construct $z \in R^{d}$ ($d=256$). It is then spatially tiled and concatenated with the visual features resulting in flattened matrix representations $\tilde{V_q} \in R^{C \times WH}$ and $\tilde{V_s} \in R^{C \times WH}$. An affinity matrix $S$ is computed to capture the similarity between them via a fully connected layer $W_{co} \in R^{C \times C}$ learning the correlation between feature channels:

\begin{equation*} 
    S = \tilde{V_s}^TW_{co}\tilde{V_q}. 
\end{equation*}
The affinity matrix $S \in R^{WH \times WH}$ relates each pixel in $\tilde{V_q}$ and $\tilde{V_s}$. A softmax operation is performed on $S$ row-wise and column-wise depending on the desired direction of relation:
\begin{align*} 
S^c = \textup{softmax}(S), \quad S^r = \textup{softmax}(S^T)
\end{align*}
For example, column $S^c_{*,j}$ contains the relevance of the $j^{th}$ spatial location in $V_q$ with respect to all spatial locations of $V_s$, where $j=1,...,WH$. The normalized affinity matrix is used to compute attention summaries $U_q$ and $U_s$:
\begin{align*}
U_q = \tilde{V_s} S^c, \quad U_s = \tilde{V_q} S^r. \quad  
\end{align*}
The attention summaries are further reshaped such that $U_q, U_s \in R^{W\times H\times C}$ and gated using a gating function $f_g$ with learnable weights $W_g$ and bias $b_g$:
\begin{align*}
f_g(U_q) &= \sigma{(W_g * U_q + b_g)}, \\ 
U_q &= f_g(U_q) \circ U_q. 
\end{align*}
Here the $\circ$ operator denotes element-wise multiplication. The gating function restrains the output to the interval $[0, 1]$ using a sigmoid activation function $\sigma$ in order to mask the attention summaries. The gated attention summaries $U_q$ are concatenated with the original visual features $V_q$ to construct the final output from the attention module to the decoder.
 
\subsection{Stacked Gated Co-Attention}


We propose to stack the multi-modal interaction module described in Section~\ref{sec:CS2QA} to learn an improved representation. Stacking allows for multiple iterations between the support and the query images. The co-attention module has two streams $f_q, f_s$ that are responsible for processing the query image and the support set images respectively. The inputs to the co-attention module, $V^i_q$ and $V^i_s$, represent the visual features at iteration $i$ for query image and support image respectively. In the first iteration, $V_q^0$ and $V_s^0$ are the output visual features from the encoder. Each multi-modal interaction then follows the recursion $\forall i=0,..,N-1$:
\begin{align*}
V_q^{i+1} &= \phi(V_q^i + f_q(V_q^i, V_s^i, z)) 
\end{align*}
The nonlinear projection $\phi$ is performed on the output from each iteration, which is composed of a 1x1 convolutional layer followed by a ReLU activation function. We use residual connections in order to improve the gradient flow and prevent vanishing gradients. The support set features $V_s^{i}, \forall i=0,..,N-1$ are computed similarly. 


\section{Temporal Object Segmentation with Few-shot Learning Setup}
We propose a novel few-shot video object segmentation (VOS) task. In this task, the image-level label for the support frame is provided to guide object segmentation in the query frames. Both instance-level and category-level setup are proposed. In the instance-level setup the query and support images are temporally related. In the category-level setup the support set and query sets are sampled from different sequences for the same object category. Even in the category-level the query set with multiple query images can be temporally related. This provides a potential research direction for ensuring the temporal stability of the learned representation that is used to segment multiple query images. 

The task is designed as a binary segmentation problem following~\cite{shaban2017one} and the categories are split into multiple folds, consistent with existing few-shot segmentation tasks defined on Pascal-$5^i$ and MS-COCO. This design ensures that the proposed task assesses the ability of few-shot video object segmentation algorithms to generalize over unseen classes. We utilize Youtube-VOS dataset training data which has 65 classes, and we split them into 5 folds. Each fold has 13 classes that are used as novel classes, while the rest are used in the meta-training phase. In the instance-level mode a randomly sampled class $Y^s$ and sequence $V = \{I_1, I_2, ..., I_N\}$ are used to construct the support set $S_p=\{(I_1, Y^s_1)\}$ and query images $I_i$. For each query image a ground-truth binary segmentation mask $M_Y^s$ is constructed by labelling all the instances belonging to $Y^s$ as foreground. Accordingly, the same image can have multiple binary segmentation masks depending on the sampled $Y^s$. During the category-level mode different sequences $V^s= \{I^s_1, I^s_2, ..., I^s_N\}$ and $V^q= \{I^q_1, I^q_2, ..., I^q_N\}$ for the same class $Y^s$ are sampled. Then random frames $\{I^s_i\}_{i=0}^{k}$ sampled from $V^s$ and $\{I^q_i\}_{i=0}^{l}$ similarly are used to construct the support and query sets respectively.

\section{Experiments}
In this section we demonstrate results of experiments conducted on the PASCAL-$5^i$ dataset~\cite{shaban2017one} compared to state of the art methods in section~\ref{sec:soa}. Not only do we set strong baselines for image level labelled few shot segmentation and outperform previously proposed work~\cite{raza2019weakly}, but we also perform close to the state of the art conventional few shot segmentation methods that use detailed pixel-wise segmentation masks. We then demonstrate the results for the different variants of our approach depicted in Fig.~\ref{fig:variants} and experiment with the proposed TOSFL setup in section~\ref{sec:ablation}.

\newcommand{\xmark}{\ding{55}}%
\newcommand{\cmark}{\ding{51}}
\begin{table*}[t!]
\centering
\begin{tabular}{|l|c|cccc|c|c||cccc|c|}
\hline
&  & \multicolumn{6}{c||}{1-shot} & \multicolumn{5}{c|}{5-shot} \\ \hline
 Method & Type & 1 & 2 & 3 & 4 & mIoU & bIoU & 1 & 2 & 3 & 4 & mIoU\\ \hline
\cite{shaban2017one} & P & 33.6 & 55.3 & 40.9 & 33.5 & 40.8 & - & 35.9 & 58.1 & 42.7 & 39.1 & 43.9\\
\cite{rakelly2018conditional} & P & 36.7 & 50.6 & 44.9 & 32.4 & 41.1 & 60.1 & 37.5 & 50.0 & 44.1 & 33.9 & 41.4\\ 
\cite{dong2018few} & P & - & - & - & - & - & 61.2 & - & - & - & - & - \\ 
\cite{siam2019amp} & P & 41.9 & 50.2 & 46.7 & 34.7 & 43.4 & 62.2 & 41.8 & 55.5 & 50.3 & 39.9 & 46.9 \\ 
\cite{wang2019panet} & P & 42.3 & 58.0 & 51.1 & 41.2 & 48.1 & 66.5 & 51.8 & 64.6 & 59.8 & 46.5 & 55.7\\ 
\cite{zhang2019canet} & P & 52.5 & 65.9 & 51.3 & 51.9 & 55.4 & 66.2 & 55.5 & 67.8 & 51.9 & 53.2 & 57.1\\ 
\cite{zhang2019pyramid} & P & 56.0 & 66.9 & 50.6 & 50.4 & 56.0 & 69.9 & 57.7 & 68.7 & 52.9 & 54.6 & 58.5 \\ \hline
\cite{zhang2019canet} & BB & - & - & - & - & \color{red}{\textbf{52.0}} & - & - & - & - & - & - \\
\cite{wang2019panet} & BB & - & - & - & - & \color{blue}{\textbf{45.1}} & - & - & - & - & - & \color{blue}{\textbf{52.8}} \\ \hline
\cite{raza2019weakly} & IL & - & - & - & - & - & \color{blue}{\textbf{58.7}} & - & - & - & - & - \\
Ours(V+S)-1 & IL & 49.5 & 65.5 & 50.0 & 49.2 & \color{red}{\textbf{53.5}} & \color{red}{\textbf{65.6}} & - & - & - & - & - \\
Ours(V+S)-2 & IL & 42.5 & 64.8 & 48.1 & 46.5 & \color{blue}{\textbf{50.5}} & \color{blue}{\textbf{64.1}} & 45.9 & 65.7 & 48.6 & 46.6 & \color{blue}{\textbf{51.7}}\\ 
 &  &  &  &  & & $\pm 0.7$ & $\pm 0.4$ &  &  &  &  & $\pm 0.07$\\ \hline
\end{tabular}
\caption{Quantitative results for 1-way, 1-shot segmentation on the PASCAL-$5^i$ dataset showing mean-Iou and binary-IoU. P: stands for using pixel-wise segmentation masks for supervision. IL: stands for using weak supervision from Image-Level labels. BB: stands for using bounding boxes for weak supervision. Red: validation scheme following~\protect\cite{zhang2019canet}. Blue: validation scheme following~\protect\cite{wang2019panet} }
\label{table:pascal5i}
\end{table*}

\subsection{Experimental Setup}
\label{sec:imp_details}
\paragraph{Network Details:} We utilize a ResNet-50~\cite{he2016deep} encoder pre-trained on ImageNet~\cite{deng2009imagenet} to extract visual features. The segmentation decoder is comprised of an iterative optimization module (IOM)~\cite{zhang2019canet} and an atrous spatial pyramid pooling (ASPP)~\cite{chen2017deeplab,chen2017rethinking}. The IOM module takes the output feature maps from the multi-modal interaction module and the previously predicted probability map in a residual form. 

\paragraph{Meta-Learning Setup:} We sample 12,000 tasks during the meta-training stage. In order to evaluate test performance, we average accuracy over 5000 tasks with support and query sets sampled from the meta-test dataset $D_{test}$ belonging to classes $L_{test}$. We perform 5 training runs with different random generator seeds and report the average of the 5 runs and the 95\% confidence interval.

\paragraph{Evaluation Protocol:} PASCAL-$5^i$ splits PASCAL-VOC 20 classes into 4 folds each having 5 classes. The mean IoU and binary IoU are the two metrics used for the evaluation process. The mIoU computes the intersection over union for all 5 classes within the fold and averages them neglecting the background. Whereas the bIoU metric proposed by~\cite{rakelly2018conditional} computes the mean of foreground and background IoU in a class agnostic manner. We have noticed some deviation in the validation schemes used in previous works. \cite{zhang2019canet} follow a procedure where the validation is performed on the test classes to save the best model, whereas~\cite{wang2019panet} do not perform validation and rather train for a fixed number of iterations. We choose the more challenging approach in~\cite{wang2019panet}.

\paragraph{Training Details:} During the meta-training, we freeze ResNet-50 encoder weights while learning both the multi-modal interaction module and the decoder. We train all models using momentum SGD with learning rate $0.01$ that is reduced by 0.1 at epoch 35, 40 and 45 and momentum 0.9. L2 regularization with a factor of 5x$10^{-4}$ is used to avoid over-fitting. Batch size of 4 and input resolution of 321$\times$321 are used during training with random horizontal flipping and random centered cropping for the support set. An input resolution of 500$\times$500 is used for the meta-testing phase similar to~\cite{shaban2017one}. In each fold the model is meta-trained for a maximum number of 50 epochs on the classes outside the test fold on pascal-$5^i$, and 20 epochs on both MS-COCO and Youtube-VOS. 

\begin{figure*}[ht!]
\begin{subfigure}{.33\textwidth}
    \centering
    \begin{subfigure}{.46\textwidth}
        \includegraphics[scale=0.3]{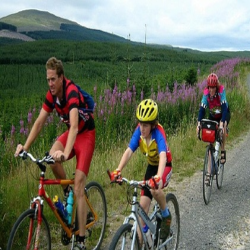}
    \end{subfigure}%
    \begin{subfigure}{.43\textwidth}
        \includegraphics[scale=0.3]{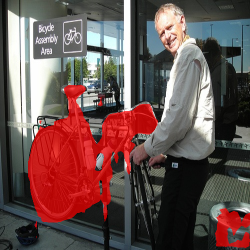}
    \end{subfigure}%
    \caption{'bicycle'}
\end{subfigure}%
\begin{subfigure}{.33\textwidth}
    \centering
    \begin{subfigure}{.46\textwidth}
        \includegraphics[scale=0.3]{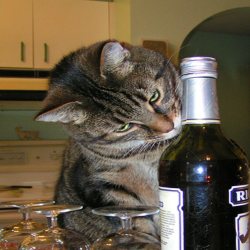}
    \end{subfigure}%
    \begin{subfigure}{.43\textwidth}
        \includegraphics[scale=0.3]{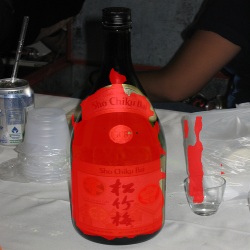}
    \end{subfigure}%
    \caption{'bottle'}
\end{subfigure}%
\begin{subfigure}{.33\textwidth}
    \centering
    \begin{subfigure}{.46\textwidth}
        \includegraphics[scale=0.3]{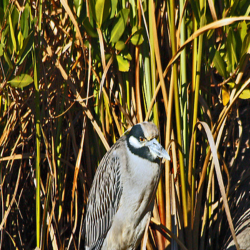}
    \end{subfigure}%
    \begin{subfigure}{.43\textwidth}
        \includegraphics[scale=0.3]{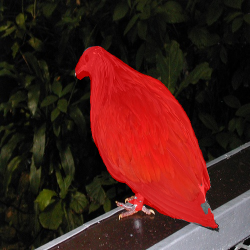}
    \end{subfigure}%
    \caption{'bird'}
\end{subfigure}

\begin{subfigure}{.33\textwidth}
    \centering
    \begin{subfigure}{.46\textwidth}
        \includegraphics[scale=0.3]{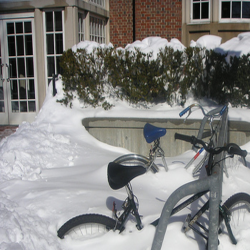}
    \end{subfigure}%
    \begin{subfigure}{.43\textwidth}
        \includegraphics[scale=0.3]{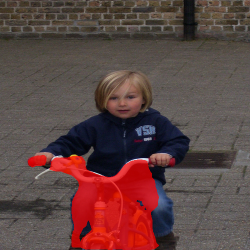}
    \end{subfigure}%
    \caption{'bicycle'}
\end{subfigure}%
\begin{subfigure}{.33\textwidth}
    \centering
    \begin{subfigure}{.46\textwidth}
        \includegraphics[scale=0.6]{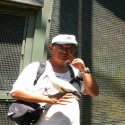}
    \end{subfigure}%
    \begin{subfigure}{.43\textwidth}
        \includegraphics[scale=0.3]{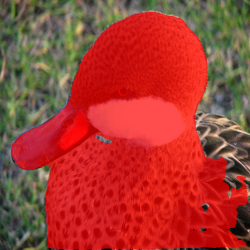}
    \end{subfigure}%
    \caption{'bird'}
\end{subfigure}%
\begin{subfigure}{.33\textwidth}
    \centering
    \begin{subfigure}{.46\textwidth}
        \includegraphics[scale=0.3]{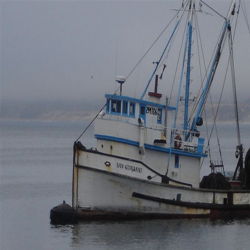}
    \end{subfigure}%
    \begin{subfigure}{.43\textwidth}
        \includegraphics[scale=0.3]{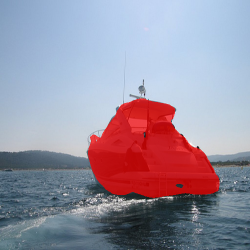}
    \end{subfigure}%
    \caption{'boat'}
\end{subfigure}
\caption{Qualitative evaluation on PASCAL-$5^i$ 1-way 1-shot. The support set and prediction on the query image are shown in pairs.}
\label{fig:pascal5i}
\end{figure*} 


\subsection{Comparison to the State-of-the-art}
\label{sec:soa}
We compare the result of our best variant (see Fig.~\ref{fig:variants}), \emph{i.e}: Stacked Co-Attention (V+S) against the other state of the art methods for 1-way 1-shot and 5-shot segmentation on PASCAL-$5^i$ in Table~\ref{table:pascal5i}. We report the results for different validation schemes. Ours(V+S)-1 follows~\cite{zhang2019canet} and Ours(V+S)-2 follows~\cite{wang2019panet}. Without the utilization of segmentation mask or even sparse annotations, our method with the least supervision of image level labels performs (53.5\%) close to the current state of the art strongly supervised methods (56.0\%) in 1-shot case and outperforms the ones that use bounding box annotations. It improves over the previously proposed image-level supervised method with a significant margin (4.8\%). For the $k$-shot extension of our method we perform average of the attention summaries during the meta-training on the $k$-shot samples from the support set. Table~\ref{table:msoco} demonstrates results on MS-COCO~\cite{lin2014microsoft} compared to the state of the art method using pixel-wise segmentation masks for the support set.

\begin{table}[t]
\centering
\begin{tabular}{|l|c|cc|}
\hline
Method & Type & 1-shot & 5-shot \\ \hline
~\cite{wang2019panet} & P & 20.9 & 29.7 \\ \hline
Ours-(V+S) & IL & 15.0 & 15.6\\ \hline
\end{tabular}
\caption{Quantitative Results on MS-COCO Few-shot 1-way.}
\label{table:msoco}
\end{table}

\subsection{Ablation Study}
\label{sec:ablation}
\begin{figure}[ht!]
    \centering
    \begin{subfigure}{.15\textwidth}
        \includegraphics[scale=0.3]{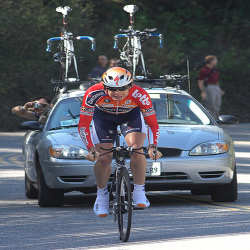}    
        \caption{Label 'Bike'}
    \end{subfigure}%
    \begin{subfigure}{.15\textwidth}
        \includegraphics[scale=0.3]{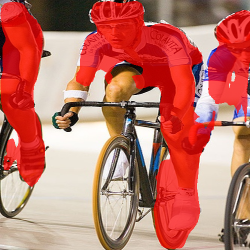}
        \caption{Prediction (V)}
    \end{subfigure}%
    \begin{subfigure}{.15\textwidth}
        \includegraphics[scale=0.3]{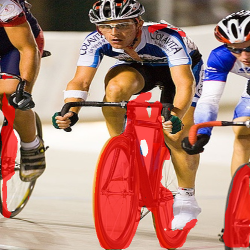}    
        \caption{Prediction (V+S)}
    \end{subfigure}
 \caption{Visual Comparison between the predictions from two variants of our method.}
    \label{fig:visual_comp}
\end{figure}

We perform an ablation study to evaluate different variants of our method depicted in Fig.~\ref{fig:variants}. Table~\ref{table:ablation_variants} shows the results on the three variants we proposed on PASCAL-$5^i$. It clearly shows that using the visual features only (V-method), lags 5\% behind utilizing word embeddings in the 1-shot case. This is mainly due to two reasons having multiple common objects between the support set and the query image and tendency to segment base classes used in meta-training. Semantic representation obviously helps to resolve the ambiguity and improves the result significantly as shown in Figure~\ref{fig:visual_comp}. Going from 1 to 5 shots, the V-method improves, because multiple shots are likely to repeatedly contain the object of interest and the associated ambiguity decreases, but still it lags behind both variants supported by semantic input. Interestingly, our results show that the baseline of conditioning on semantic representation is a very competitive variant: in the 1-shot case it even outperforms the (V+S) variant. However, the bottleneck in using the simple scheme to integrate semantic representation depicted in Fig.~\ref{fig:variants:semantic} is that it is not able to benefit from multiple shots in the support set. The (V+S)-method in the 5-shot case improves over the 1-shot case by 1.2\% on average over the 5 runs, which confirms its ability to effectively utilize more abundant visual features in the 5-shot case. One reason could explain the strong performance of the (S) variant. In the case of a single shot, the word embedding pretrained on a massive text database may provide a more reliable guidance signal than a single image containing multiple objects that does not necessarily have visual features close to the object in the query image.

\begin{table}[]
\centering
\begin{tabular}{|l|c|c|}
\hline
Method & 1-shot & 5-shot\\ \hline
V & $44.4 \pm 0.3$ & $49.1 \pm 0.3$ \\ \hline
S & $\textbf{51.2} \pm 0.6$ & $ 51.4 \pm 0.3$ \\ \hline
V+S & $50.5 \pm 0.7$ & $\textbf{51.7} \pm 0.07$ \\ \hline
\end{tabular}
\caption{Ablation Study on 4 folds of Pascal-$5^i$ for few-shot segmentation for different variants showing mean-IoU. V: visual, S: semantic. V+S: both features.}
\label{table:ablation_variants}
\end{table}

\begin{table}[t]
\centering
\begin{tabular}{|l|ccccc|c|}
\hline
Method & 1 & 2 & 3 & 4 & 5 & Mean-IoU \\ \hline
V & 40.8 & 34.0 & 44.4 & 35.0 & 35.5 & $38.0 \pm 0.7$ \\ 
S & 42.7 & 40.8 & 48.7 & 38.8 & 37.6 & $41.7 \pm 0.7$ \\ 
V+S & \textbf{46.1} & \textbf{42.0} & \textbf{50.7} & \textbf{41.2} & \textbf{39.2} & $\textbf{43.8} \pm 0.5$ \\ \hline
\end{tabular}
\caption{Ablation Study on 5 folds on Youtube-VOS Instance-level TOSFL. V: visual, S: semantic. V+S: both features.}
\label{table:ytbvos}
\end{table}

Table~\ref{table:ytbvos} shows the results on our proposed novel video object segmentation instance-level task, comparing variants of the proposed approach. As previously, the baseline V-method based on co-attention module with no word embeddings, lags behind both S- and (V+S)-methods. It is worth noting that unlike the conventional video object segmentation setups, the proposed video object segmentation task poses the problem as a binary segmentation task conditioned on the image-level label. Both support and query frames can have multiple salient objects appearing in them, however the algorithm has to segment only one of them corresponding to the image-level label provided in the support frame. According to our observations, this multi-object situation occurs in this task much more frequently than \emph{e.g.} in the case of Pascal-$5^i$. Additionally, not only the target, but all the nuisance objects present in the video sequence will relate via different viewpoints or deformations. We demonstrate in Table~\ref{table:ytbvos} that the (V+S)-method's joint visual and semantic processing in such scenario clearly provides significant gain. 

\section{Conclusion}
In this paper we proposed a multi-modal interaction module that relates the support set image and query image using both visual and word embeddings. We proposed to meta-learn a stacked co-attention module that guides the segmentation of the query based on the support set features and vice versa. The two main takeaways from the experiments are that (i) few-shot segmentation significantly benefits from utilizing word embeddings and (ii) it is viable to perform high quality few-shot segmentation using stacked joint visual semantic processing with weak image-level labels.

\appendix
\section{Additional Results}
\begin{figure}[ht!]
    \centering
    
    \begin{subfigure}{.15\textwidth}
        \includegraphics[scale=0.15]{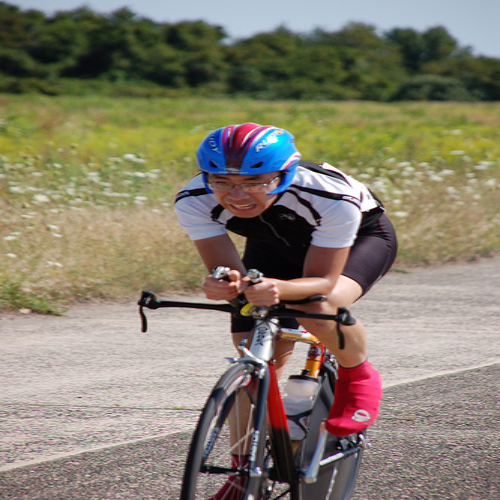}    
    \end{subfigure}%
    \begin{subfigure}{.15\textwidth}
        \includegraphics[scale=0.15]{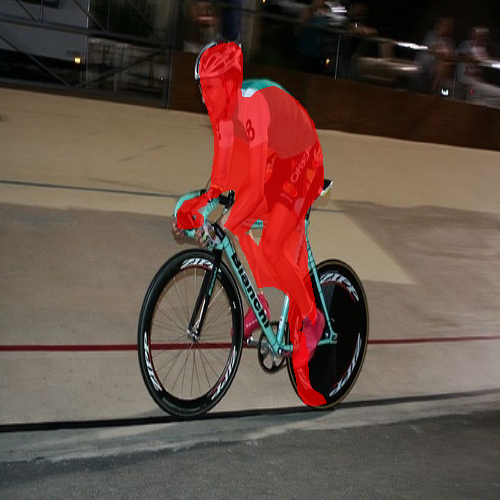}
    \end{subfigure}%
    \begin{subfigure}{.15\textwidth}
        \includegraphics[scale=0.15]{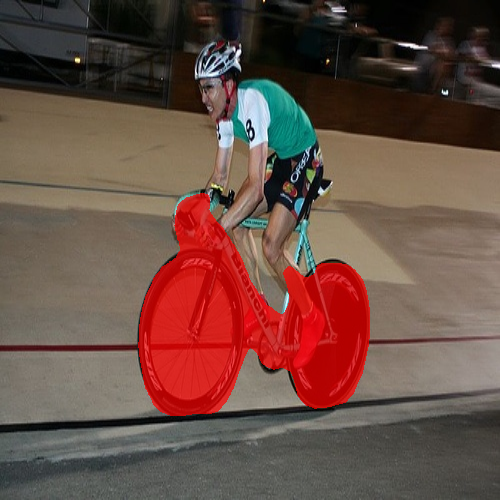}    
    \end{subfigure}

    \begin{subfigure}{.15\textwidth}
        \includegraphics[scale=0.15]{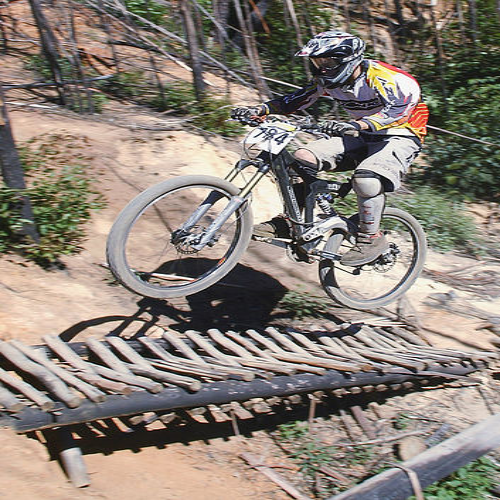}    
    \end{subfigure}%
    \begin{subfigure}{.15\textwidth}
        \includegraphics[scale=0.15]{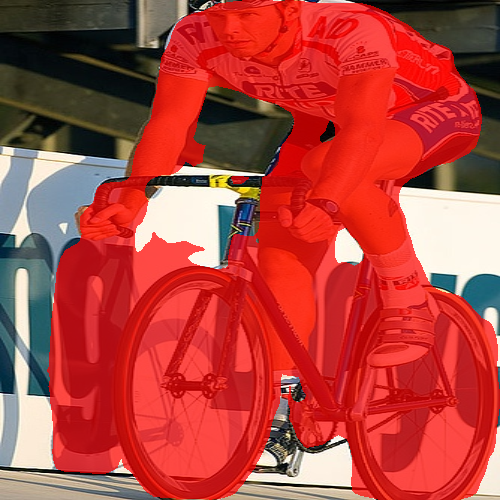}
    \end{subfigure}%
    \begin{subfigure}{.15\textwidth}
        \includegraphics[scale=0.15]{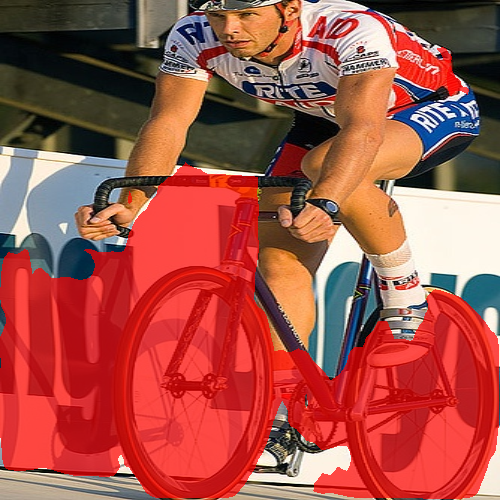}    
    \end{subfigure}
    
    \begin{subfigure}{.15\textwidth}
        \includegraphics[scale=0.15]{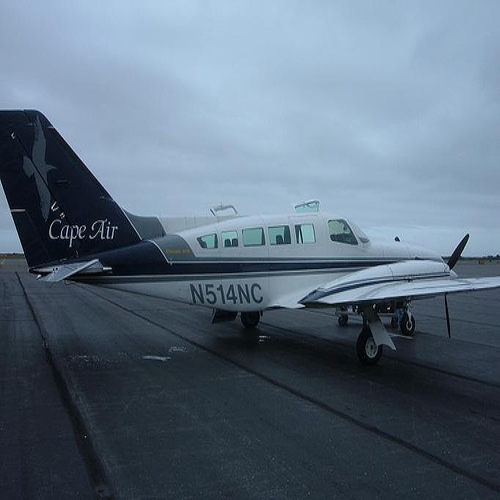}    
    \end{subfigure}%
    \begin{subfigure}{.15\textwidth}
        \includegraphics[scale=0.15]{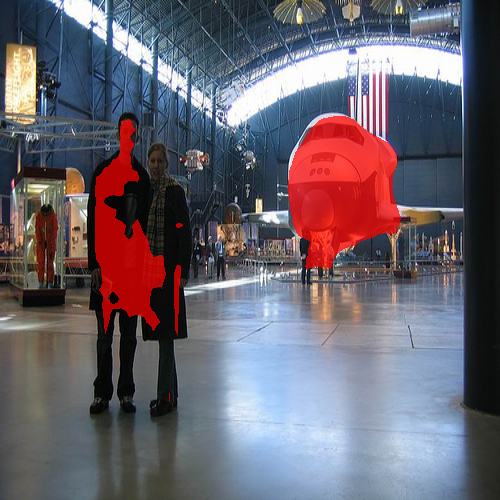}
    \end{subfigure}%
    \begin{subfigure}{.15\textwidth}
        \includegraphics[scale=0.15]{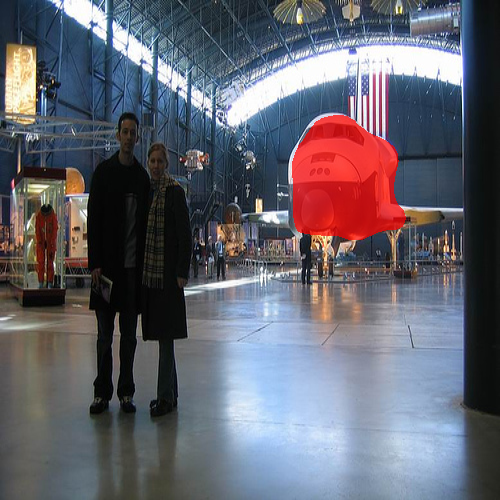}    
    \end{subfigure}
    
    \begin{subfigure}{.15\textwidth}
        \includegraphics[scale=0.15]{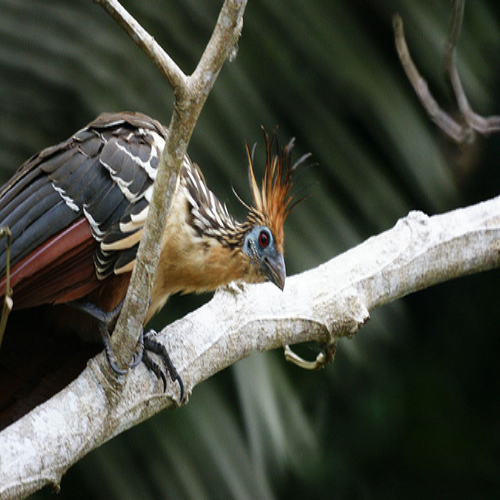}
        \caption{Support Set}
    \end{subfigure}%
    \begin{subfigure}{.15\textwidth}
        \includegraphics[scale=0.15]{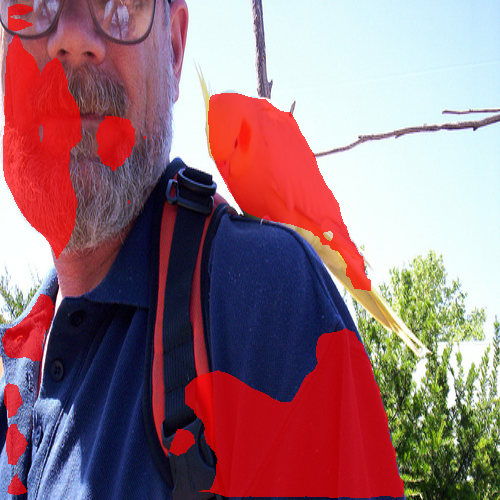}
        \caption{Ours(V)}
    \end{subfigure}%
    \begin{subfigure}{.15\textwidth}
        \includegraphics[scale=0.15]{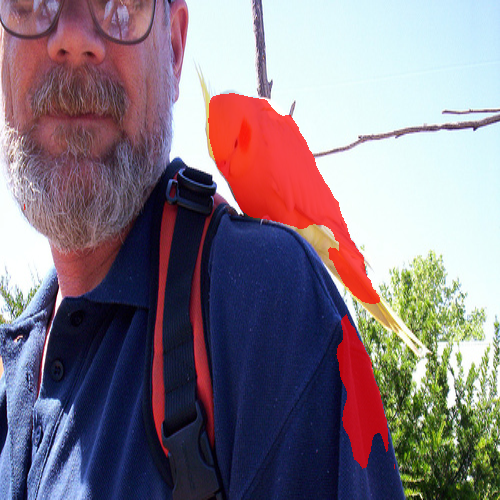}    
        \caption{Ours (V+S)}
    \end{subfigure}
    \caption{Qualitative analysis on fold 0 pascal-$5^i$ between our method (V+S) and ours (V) that can not disambiguate multiple common objects and is biased toward base classes used in training.}
    \label{fig:qual2}
\end{figure}

In this section we present additional qualitative and quantitative results on pascal-$5^i$ and Youtube-VOS datasets.
We show more qualitative results on pascal-$5^i$ that motivate the benefit of using both visual and semantic features in Fig.~\ref{fig:qual2}. It shows in first two rows that when both the query and support set have multiple common objects semantic features help to disambiguate this situation. It also shows in the last two rows that with visual features only there is a higher tendency to segment partly pixels belonging to classes that were provided in the meta-training stage as well. In this case semantic features again help to disambiguate this situation. 

We further show another ablation study to evaluate different components in Tables~\ref{table:ablation_components_pascal, table:ablation_components_pascal}. We compare using a simple conditioning on the support set features through concatenation with the query visual features against performing co-attention between support and query feature maps. It shows clearly the benefit from performing co-attention. Nonetheless, visual features solely is not capable to disambiguate the above mentioned situations, while the visual with semantic features even with simple concatenation shows an improvement. Further combining semantic features with stacked co-attention shows further gain on both pascal-$5^i$ and Youtube-VOS. In Table~\ref{table:categorytosfl} we compare our variants in category-level TOSFL setup, in which the support set is sampled from a different sequence than the query set. It shows similar conclusions to Pascal-$5^i$ results.

\begin{figure}[ht!]
\centering
\begin{subfigure}{.15\textwidth}
        \includegraphics[scale=0.3]{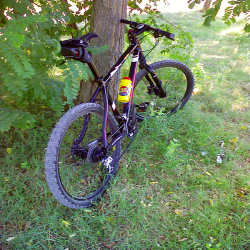}
\end{subfigure}%
\begin{subfigure}{.14\textwidth}
        \includegraphics[scale=0.3]{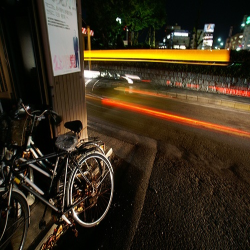}
\end{subfigure}%
\begin{subfigure}{.14\textwidth}
        \includegraphics[scale=0.41]{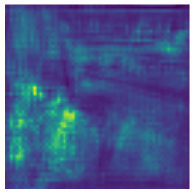}
\end{subfigure}
%
\caption{Visualizing Output from Gated Co-Attention on PASCAL-$5^i$ query images. (a) Support Set Image. (b) Query Set Image. (c) Gated Attention Map Output.}
\label{fig:attnetion}
\end{figure}

Finally we visualize the output gated attention maps in Fig.~\ref{fig:attnetion} to show the intermediate output from performing co-attention with both visual and semantic features. It demonstrates that our multi-modal interaction module in the first row successfully attends to the pixels belonging to the novel class.

\begin{table}[t!]
\centering
\begin{tabular}{|l|c|}
\hline
Method & mIoU \\ \hline
V-Cond & 42.7 \\ 
V-CoAtt & 44.6 \\ \hline
V+S-Cond & 50.1 \\ 
V+S-CoAtt & 50.2 \\ 
V+S-SCoAtt & \textbf{51.0} \\ \hline
\end{tabular}
\caption{Ablation Study for different components with 1 run on Pascal-$5^i$. V: visual, S: semantic. SCoAtt: Stack Co-Attention. Cond: Concatenation based conditioning.}
\label{table:ablation_components_pascal}
\end{table}

\begin{table}[t!]
\centering
\begin{tabular}{|l|c|}
\hline
Method & mIoU \\ \hline
V+S-Cond &  42.3\\
V+S-SCoAtt & \textbf{43.7}\\ \hline
\end{tabular}
\caption{Ablation Study for different components with 1 run on Youtube-VOS. V: visual, S: semantic. SCoAtt: Stack Co-Attention. Cond: Concatenation based conditioning.}
\label{table:ablation_components_ytbvos}
\end{table}

\begin{table}[t!]
\centering
\begin{tabular}{|l|c|}
\hline
Method & mIoU \\ \hline
V-CoAtt & 36.1 \\
S-Cond & \textbf{37.7} \\
V+S-SCoAtt & 37.6 \\  \hline
\end{tabular}
\caption{Ablation study for different variants on the Category Level TOSFL 1-shot on Youtube-VOS.}
\label{table:categorytosfl}
\end{table}

\bibliographystyle{named}
\bibliography{ijcai20}

\end{document}